# An Adaptive Synaptic Array using Fowler-Nordheim Dynamic Analog Memory

Darshit Mehta[1], Kenji Aono[2] and Shantanu Chakrabartty[1,2,*]

## Abstract

*In this paper we present a synaptic array that uses dynamical states to implement an analog memory for energy-efficient training of machine learning (ML) systems. Each of the analog memory elements is a micro-dynamical system that is driven by the physics of Fowler-Nordheim (FN) quantum tunneling, whereas the system level learning modulates the state trajectory of the memory ensembles towards the optimal solution. We show that the extrinsic energy required for modulation can be matched to the dynamics of learning and weight decay leading to a significant reduction in the energy-dissipated during ML training. With the energy-dissipation as low as 5 fJ per memory update and a programming resolution up to 14 bits, the proposed synapse array could be used to address the energy-efficiency imbalance between the training and the inference phases observed in artificial intelligence (AI) systems.*

## Introduction

Implementation of reliable and scalable synaptic weights or memory remains an unresolved challenge in the design of energy-efficient machine learning (ML) and neuromorphic processors [1]. Ideally, the synaptic weights should be "analog" and should be implemented on a non-volatile, easily modifiable storage device [2]. Furthermore, if these memory elements are integrated in proximity with the computing circuits or processing elements, then the resulting compute-in-memory (CIM) architecture [3, 4] has the potential to mitigate the "memory wall" [5, 6, 7] which refers to the energy-efficiency bottleneck in ML processors that arises due to repeated memory access. In most practical and scalable implementations, the processing elements are implemented using CMOS circuits; as a result, it is desirable that the analog synaptic weights be implemented using a CMOS-compatible technology. In literature, several multi-level non-volatile memory devices have been proposed for implementing analog synapses. These include the cross-bar memristor based resistive random-access memories (RRAM) [8], magnetic random-access memories (MRAM) [9], Phase Change Memory (PCM) [10], Spin Torque Transfer RAM (STTRAM) [11], Conductive Bridge RAM [12] or the three terminal devices like the floating-gate transistors [13], ferroelectric field-effect transistor-based RAM (FeRAM) [14], Charge Trap Memory [15] and Electrochemical RAMs (ECRAM) [16]. In all these devices the analog memory states are static in nature, where each state needs to be separated from others by an energy barrier $\Delta E$. In non-volatile storage, it is critical that this energy-barrier is chosen to be large enough to prevent memory leakage due to thermal-fluctuations or other environmental disturbances. For example, in memristive devices the state of the conductive filament between two electrodes determines the stored analog value, whereas in charge-based devices like floating-gates or FeFET, the state of polarization determines the analog value. At a fundamental level, the energy dissipated to transition between different analog states is determined by the energy-barrier $\Delta E$. For example, switching the RRAM memory state requires 100 fJ per bit [17], whereas STT_MRAM requires about 4.5pJ per bit [18]. A learning/training algorithm that adapts its weights in quantized steps

[1]Department of Biomedical Engineering, Washington University in St. Louis
[2]Department of Electrical and Systems Engineering, Washington University in St. Louis
[*]Corresponding author: shantanu@wustl.edu

(…, $W_{n-1}$, $W_n$, $W_{n+1}$, …) towards a target solution (or local extrema), must dissipate energy (…, $\Delta E_{n-1}$, $\Delta E_n$, $\Delta E_{n+1}$, …) for memory updates, as shown in Fig. 1(a).

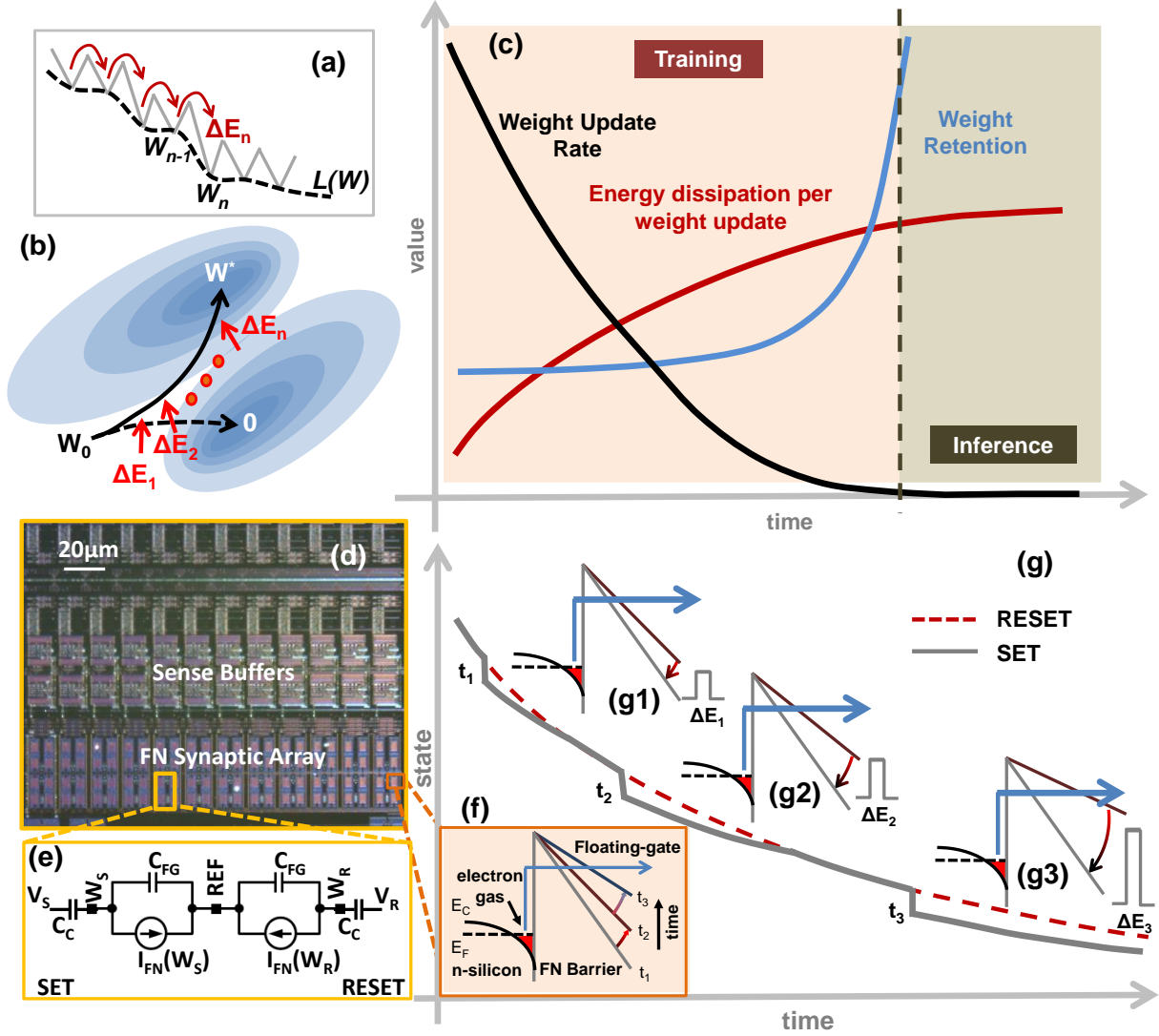

*Figure 1: Motivation and principle of operation for the proposed synaptic memory device: (a) conventional non-volatile analog memory where transition between analog static states dissipates energy; (b) Dynamic analog memory where an external energy is used to modulate the trajectory of the memory states towards the optimal solution; (c) desired analog synapse characteristic where the memory retention rate is traded-off with the write energy; reducing the energy dissipation per weight update in training phase by matching the dynamics of the dynamic analog memory to the weight decay; (d) micrograph of a fabricated DAM array along with (e) its equivalent circuit where the leakage current $I_{FN}$ is implemented by (f) the electron transport across a Fowler-Nordheim (FN) tunneling barrier; (g) Implementation of the FN tunneling based DAM where dynamic states g1-g3 determines the energy dissipated per memory update and memory retention rate.*

In this paper we present a synaptic memory device that uses dynamical states (instead of static states) to implement analog memory in an effort to improve the energy-efficiency of ML training. The core of the proposed device is itself a micro-dynamical system and the system-level learning/training process modulates the dynamical state (or state trajectory) of the memory ensembles. The concept is illustrated in Fig. 1(b), which shows a reference ensemble trajectory that continuously decays towards a *zero vector* without the presence of any external modulation. However, during the process of learning, the trajectory of the memory ensemble is pushed towards an optimal solution **W\***. The main premise of this paper is that the extrinsic energy (… *ΔE$_{n-1}$, ΔE$_n$, ΔE$_{n+1}$, …*) required for modulation, if matched to the dynamics of learning, could reduce the energy-budget for ML training. This is illustrated in Fig. 1(c) which shows a convergence plot corresponding to a typical ML system as it transitions from a training phase to an inference phase. During the training phase, the synaptic weights are adapted based on some learning criterion whereas in the inference phase the synaptic weights remain fixed or are adapted intermittently to account for changes in the operating conditions. Generally, during the training phase the amount of weight updates is significantly higher than in the inference phase, as a result, memory update operations require a significant amount of energy. Take for example support-vector machine (SVM) training, the number of weight updates scale quadratically with the number of support vectors and the size of the training data, whereas adapting the SVM during inference only scales linearly with the number of support-vectors [19]. Thus, for a constant energy dissipation per update, the total energy-dissipated due to weight updates is significantly higher in training than during inference. However, if the energy-budget per weight updates could follow a temporal profile as shown in Fig.1c, wherein the energy dissipation is no longer constant, but inversely proportional to the expected weight update rate, then the total energy dissipated during training could be significantly reduced. One way to reduce the weight update or memory write energy budget is to trade-off the weight's retention rate according to the profile shown in Fig. 1c. During the training phase, the synaptic element can tolerate lower retention rates or parameter leakage because this physical process could be matched to the process of weight decay or regularization, techniques commonly used in ML algorithms to achieve better generalization performance [20]. As shown in Fig. 1c, the memory's retention rate should increase as the training progresses such that at convergence or in the inference phase the weights are stored on a non-volatile memory.

In this paper we describe a dynamic analog memory (DAM) that can exhibit a temporal profile similar to that of Fig. 1c. Furthermore, the memory is implemented on a standard CMOS process without the need for any additional processing layers. Fig. 1e shows a micrograph of a DAM array and in the Supplementary Section I we describe the circuit implementation details. The proposed DAM requires a Fowler-Nordheim (FN) quantum-tunneling barrier which can be created by injecting sufficient electrons onto a polysilicon island (floating-gate) that is electrically isolated by thin silicon-di-oxide barriers [21]. As the electron tunnels through the triangular barrier, as shown in Fig. 1f, the barrier profile changes which further inhibits the tunneling of electrons. We have previously shown that the dynamics of this simple system is robust enough to implement time-keeping devices [22] and self-powered sensors [23]. In this paper, we use a pair of synchronized FN-dynamical systems to implement a DAM suitable for implementing ML training/inference engines. Figure 1(f) shows the dynamics of two FN-dynamical systems, labeled as SET and RESET, whose analog states continuously and synchronously decay with respect to time. In our previous work [23], we have shown the dynamics across different FN-dynamical systems can be synchronized with respect to each other with an accuracy greater than 99.9%. However, when an external voltage pulse modulates the SET system, as shown in Fig. 1f, the dynamics of the SET system becomes desynchronized with respect to the RESET system. The degree of desynchronization is a function of the state of the memory at different time instances (Fig. 1g, insets g1-g3) which determines the memory's retention rate. For instance, at time-instant $t_1$, a small magnitude pulse would produce the same degree of desynchronization as a large magnitude pulse at the time-instant $t_3$. However, at $t_1$ the pair of desynchronized systems (SET and RESET) would resynchronize more rapidly as compared to desynchronized systems at time-instants $t_2$ or $t_3$. This resynchronization effect results in shorter data retention; however, this feature could be leveraged to implement weight-decay in ML training. At time-instant $t_3$, the resynchronization effect is weak enough that the FN-dynamical system acts as a persistent non-volatile memory with high data-retention time. In Methods section, we derive the FN-dynamical system mathematical model and compare it to ML training formulation. We show that the energy required for updating the memory and its data retention capacity can be annealed according to the profile shown in Fig. 1c.

## Results

### Dynamic analog memory with an asymptotic non-volatile storage

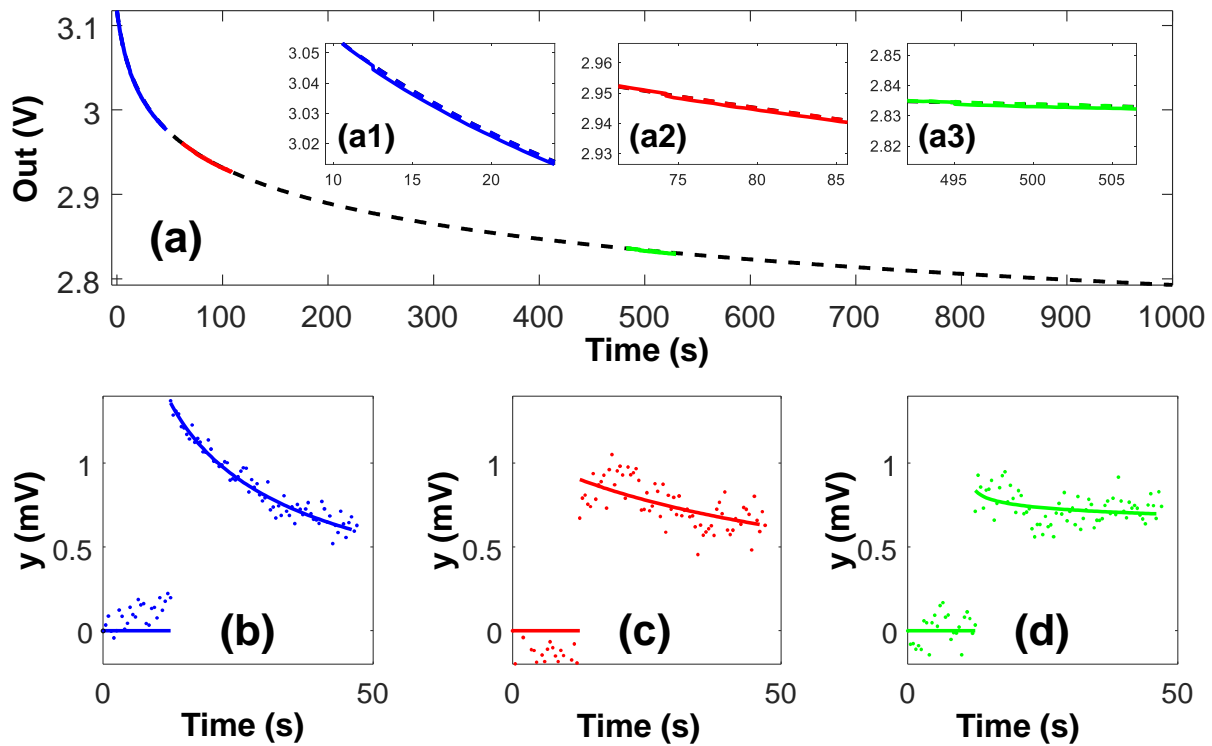

*Figure 2: a) $W_S$ (solid line) and $W_R$ (dashed line) response at 3 different operating conditions (zoomed insets: a1, a2, a3). b-d) DAM response calculated from the $W_S$ and $W_R$ voltage values.*

The dynamics of the FN-tunneling based DAM (or FN-DAM) were verified using prototypes fabricated in a standard CMOS process (micrograph shown in Fig. 1e.). The FN-DAM devices were programed and initialized through a combination of FN tunneling and hot electron injection. Detailed description of the general programming process can be found in [23] with implementation specific notes in the Methods section. The tunneling nodes ($W_S$ and $W_R$ in Fig. 1e) were initialized to around 8 V and decoupled from the readout node by a decoupling capacitor to the sense buffers (shown in supplementary Fig. 1). The readout nodes were biased at a lower voltage (~3 V) to prevent hot electron injection [24] onto the floating gate during readout operation. Fig. 2 shows the measured dynamics of the FN-DAM device in different initialization regimes used in ML training, as described in Fig. 1c. The different regimes were obtained by initializing the tunneling nodes ($W_S$ and $W_R$) to different voltages (see Methods section), whilst ensuring that the tunneling rates on the $W_S$ and $W_R$ nodes were equal. Initially (during the training phase), tunneling-node voltages were biased high (readout node voltage of 3.1 V), leading to faster FN tunneling (Fig. 2, inset a). A square input pulse of 100 mV magnitude and 500 ms duration (5 fJ of energy) was found to be sufficient to desynchronize the SET node by 1 mV. However, as shown in Fig. 2(b), the rate of resynchronization in this regime is high leading to a decay in the stored weight down to 30% in 40 s. At t = 90 s, the voltage at node $W_S$ has reduced (readout node voltage of 2.9 V), and a larger voltage amplitude (500 mV) is required to achieve the same desynchronization magnitude of 1 mV, corresponding to an energy expenditure of 125 fJ. However, as shown in Fig. 2(c), the rate of resynchronization is low in this regime, leading to a decay in the stored weight down to 70% its value in 40 s. Similarly, at a later time instant t = 540 s, a 1 V signal desynchronizes the recorder by 1 mV, however as shown in Fig. 1(d), in this regime 95% of the stored weight value is retained after 40 s. This mode of operation is suitable during the inference phase of machine learning when the weights have already been trained, but the models need to be sporadically adapted to account for statistical drifts. Modeling studies described in Supplementary Section II shows that the write energy per update starts from as low as 5 fJ and increases to 2.5 pJ over a period of a period of 12 days. Supplementary Fig. 3 indicates that at lower $W_S$/$W_R$ operating voltage (~ 6V) or at greater instants of time the retention time of FN-DAM converges to that of other FLASH based memory.

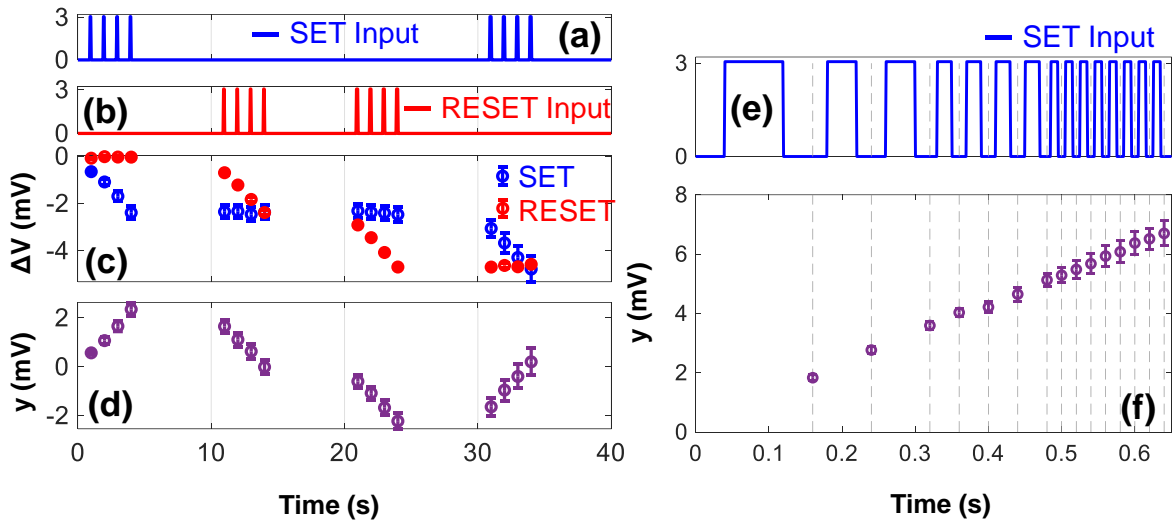

*Figure 3: (a-b) FN-DAM response to SET pulses of varying frequency. c) Change in $W_S$ and $W_R$ potentials due to SET and RESET pulses. g) DAM response calculated as difference between $W_S$ and $W_R$ voltages. Error bars indicate standard deviation estimated across 12 devices.*

Each DAM in the FN-DAM device was programmed by independently modulating the SET and RESET junctions shown in Fig. 1(e). The corresponding $W_S$ and $W_R$ nodes were initially synchronized with respect to each other. After a programming pulse was applied to the SET or RESET control gate, the difference between the voltages at the $W_S$ and $W_R$ nodes were measured using an array of sense buffers. In results shown in Fig. 3a-d, a sequence of 100 ms SET and RESET pulses were applied. The measured difference between the voltages at the $W_S$ and $W_R$ nodes indicates the state of the memory. Each SET pulse increases the state while a RESET pulse decreases the state. In this way, the FN-device can implement a DAM that is bidirectionally programmable with unipolar pulses. Fig. 3d also shows the cumulative nature of the FN-DAM updates which implies that the device can work as an incremental/decremental counter.

Fig. 3e-f show measurement results which demonstrate the resolution at which a FN-DAM can be programmed as an analog memory. The analog state can be updated by applying digital pulses of varying frequency and variable number of pulses. In Fig. 3e, four cases of applying a 3 V SET signal for a total of 100 ms are shown: a single 100 ms pulse; two 50 ms pulses; four 25 ms pulses; and eight 12.5 ms pulses. The results show the net change in the stored weight was consistent across the 4 cases. A higher frequency leads to a finer control of the analog memory updates. Note that any variations across the devices can be calibrated or mitigated by using an appropriate learning algorithm [25]. The variations could also be reduced by using careful layout techniques and precise timing of the control signals.

## Characterization of FN-DAM

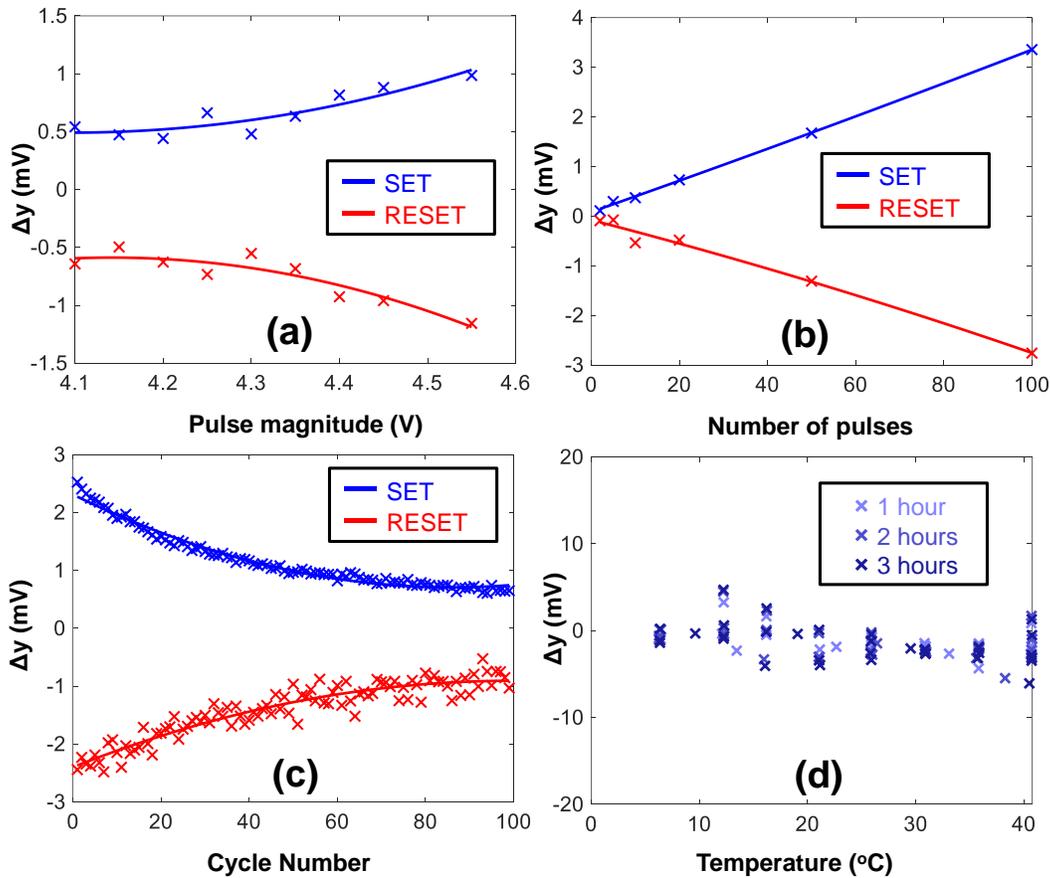

*Figure 4: Device characterization: (a) Change in DAM response with each pulse of same magnitude and duration. b) DAM response to varying number of pulses. c) DAM response to pulses of different magnitude. d) device state drift due to temperature variations after 1,2 and 3 hours.*

The FN-DAM device can be programmed by changing the magnitude of the SET/RESET pulse or its duration (equivalently number of pulses of fixed duration). Fig. 4a shows response when the magnitude of the SET and RESET input signals varies from 4.1 V to 4.5 V. The measured response shown in Fig. 4a shows an exponential relationship with the amplitude of the signal. When short-duration pulses are used for programming, the stored value varies linearly with the number of pulses, as shown in Fig. 4b. However, repeated application of pulses with constant magnitude produces successively smaller change in programmed value due to the dynamics of the DAM device (Fig. 4a). One way to achieve a constant response is to pre-compensate the SET/RESET control voltages such that a target voltage difference $y = (W_S - W_R)$ can be realized. The differential architecture increases the device state robustness against disruptions from thermal fluctuations (Fig. 4d). The stored value on DAM devices will leak due to thermal-induced processes or due to trap-assisted tunneling. However, in DAM, the weight is stored as difference in the voltages corresponding to $W_S$ and $W_R$ tunneling junctions which are similarly affected by temperature fluctuations. To verify this, we exposed the FN-DAM device to temperature ranging from 5 – 40 °C. Fig. 4d shows that the DAM response is robust to temperature variation and the amount of desynchronization for a single recorder never exceeds 20 mV. When responses from multiple FN-DAM devices are pooled together, the variation due to temperature further reduces.

## FN-DAM based Classifier

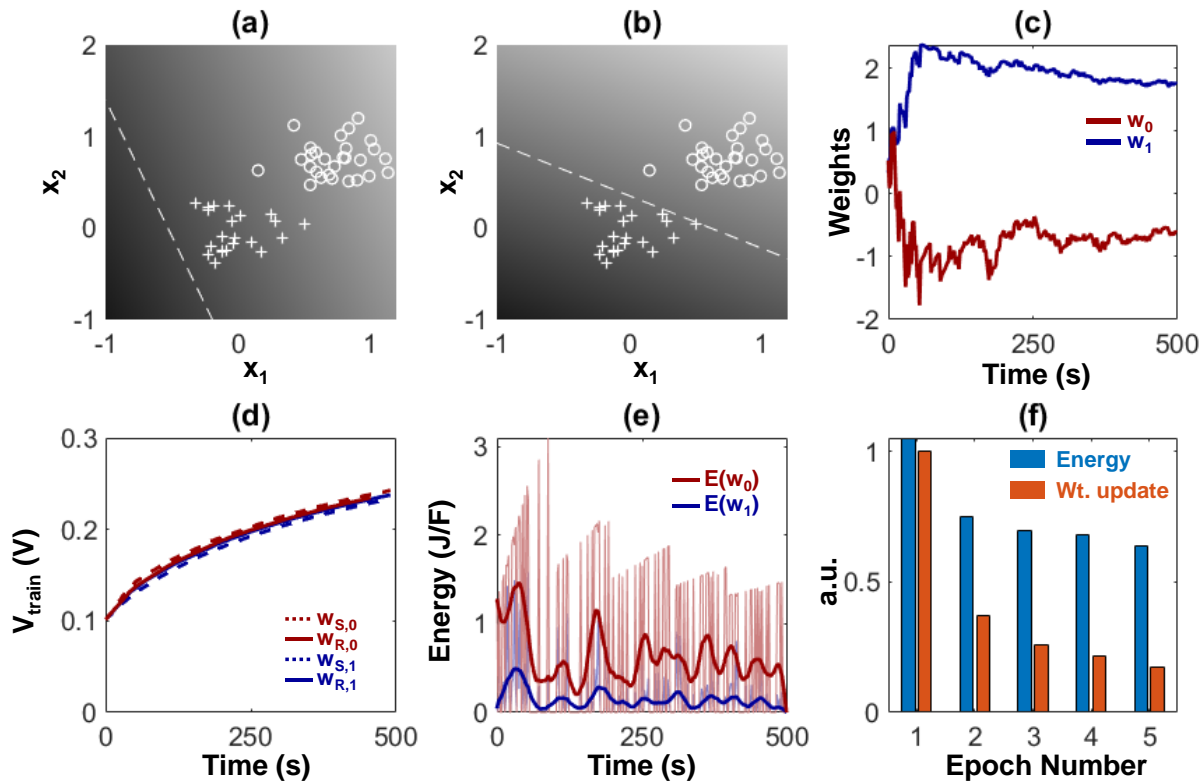

*Figure 5: Synaptic memory for neuromorphic applications a) Test data set with randomly initialized decision boundary b) Decision boundary after training. c) Evolution of weights after 5 epochs. d) Input voltage required for initiating a unit change in weight. e) Energy expended in updating the weights. f) Average magnitude of weight update and average energy required for each epoch.*

In this section we experimentally demonstrate the benefits of exploiting the dynamics of FN-DAM weights when training a simple linear classifier. For this results, two FN-DAM devices were independently programmed according to the perceptron training rule [26]. We trained the weights of a perceptron model to classify a linearly separable dataset comprises 50 instances of two-dimensional vectors, shown in Fig. 5a. During each epoch, the network loss function and gradients were evaluated for every training point in a randomized order, with time interval between successive training points being two seconds. Fig. 5b shows that after training for 5 epochs, the learned boundary can correctly classify the given data. Fig. 5c shows the evolution of weights as a function of time. As can be noted in the figure, initially the magnitude of weight updates (negative of the cost function gradient) was high for the first 50 seconds, after which the weights stabilized and required smaller updates. The energy consumption of the training algorithm can be estimated based on the magnitude and number of the SET/RESET pulses required to carry out the required update for each misclassified point. As the SET/RESET nodes evolve in time, they require larger voltages for carrying out updates, shown in Fig. 5d. The gradient magnitude was mapped onto an equivalent number of 1 kHz pulses, rounding to the nearest integer. Fig. 5e shows the energy (per unit capacitance) required to carry out the weight update whenever a point was misclassified. Though the total magnitude of weight update decreased with each epoch, the energy required to carry out the updates had lower variation (Fig. 5f). The relatively larger energy required for smaller weight updates at later epochs led to longer retention times of the weights (Supplementary Fig. 3).

## Discussions

In this paper we reported a Fowler-Nordheim quantum tunneling based dynamic analog memory (FN-DAM) whose physical dynamics can be matched to the dynamics of weight updates used in machine learning (ML) or neural network training. During the training phase, the weights stored on FN-DAM are plastic in nature and decay according to a learning-rate evolution that is necessary for the convergence of gradient-descent training [27]. As the training phase transitions to an inference phase, the FN-DAM acts as a non-volatile memory. As a result, the trained weights are persistently stored without requiring any additional refresh steps (used in volatile embedded DRAM architectures [28]). The plasticity of FN-DAM during the training phase can be traded off with the energy-required to update the weights. This is important because the number of weight updates during training scale quadratically with the number of parameters, hence the energy-budget during training is significantly higher than the energy-budget for inference. The dynamics of FN-DAM bears similarity to the process of annealing used in neural network training and other stochastic optimization engines to overcome local minima artifacts [29]. Thus, it is possible that FN-DAM implementations or ML processors can naturally implement annealing without dissipating any additional energy. If such dynamics were to be emulated on other analog memories, it would require additional hardware and control circuitry. In the Supplementary Section IV, we show that an FN-DAM based deep neural network (DNN) can achieve similar classification accuracy as a conventional DNN while dissipating significantly less energy during training. Note that for this demonstration, only the fully connected layers were trained while the feature layers were kept static. This mode of training is common for many practical DNN implementations on edge computing platforms where the goal is not only to improve the energy-efficiency of inference but also for training [30].

Several challenges exist in scaling the FN-DAM to large neural-networks. Training a large-scale neural network could take days to months [31] depending on the complexity of the problem, complexity of the network, and the size of the training data. This implies that the FN-DAM dynamics need to match the long training durations as well. Fortunately, the *1/log* characteristics of FN devices ensures that the dynamics could last for durations greater than a year [32] The other challenge that might limit the scaling of FN-DAM to large neural network is the measurement precision. The resolution of the measurement and the read-out circuits limit the energy-dissipated during memory access and how fast the gradients can be computed (Supplementary Fig. 5). For instance, a 1 pF floating-gate capacitance can be initialized to store $10^7$ electrons. Even if one were able to measure the change in synaptic weights for every electron tunneling event, the read-out circuits would need to discriminate 100 nV changes. A more realistic scenario would be measuring the change in voltage after 1000 electron tunneling events which would imply measuring 100 µV changes. However, this will reduce the resolution of the stored weights/updates to 14 bits. This resolution might be sufficient for training a medium sized neural network; however, it is still an open question if this resolution would be sufficient for training large-scale networks [33, 34]. A mechanism to improve the dynamic range and the measurement resolution is to use a current-mode readout integrated with current-mode neural network architecture. If the read-out transistor is biased in weak-inversion, 120 dB of dynamic range could be potentially achieved. However, note that even in this operating mode, the resolution of the weight would still be limited by the number of electrons and the quantization due to electron transport. Addressing this limitation would be a part of future research.

Another limitation that arises due to finite number of electrons stored on the floating-gate and transported across the tunneling barrier during SET and RESET, is the speed of programming. Shorter duration programming pulses would reduce the change in stored voltage (weight) which could be beneficial if precision in updates is desired. In contrast, by increasing the magnitude of the programming pulses, as shown in Fig. 4(a), the change in stored voltage can be coarsely adjusted. However, this would limit the number of updates before the weights saturate. Note that due to device mismatch the programmed values would be different on different FN-DAM devices.

In terms of endurance, after a single initialization the FN-DAM can support $10^3$–$10^4$ update cycles before the weight saturates. However, at the core FN-DAM is a FLASH technology and could potentially be reinitialized again. Given that the endurance of FLASH memory is $10^3$ [35], it is anticipated that FN-DAM to have an endurance of $10^6$–$10^7$ cycles. In terms of other memory performance metrics, the $I_{ON}/I_{OFF}$ ratio for the FN-DAM is determined by the operating regime and the read-out mechanism. Supplementary Fig. 6 shows the expected ratio estimated using the FN-DAM model. Also, FN-DAM when biased as a non-volatile memory requires on-chip charge-pumps only to generate high-voltage programming pulses for infrequent global erase; thus, compared to FLASH memory, FN-DAM should have fewer failure modes [36].

The main advantage of FN-DAM compared to other emerging memory technologies is its scalability and compatibility with CMOS. At its core, FN-DAM is based on floating-gate memories which have been extensively studied in context of machine learning architectures [13]. Furthermore, from an equivalent circuit point of view, FN-DAM could be viewed as a capacitor whose charge can be precisely programmed using CMOS processing elements. FN-DAM also provides a balance between weight-updates that are not too small so that learning never occurs versus weight-updates being too large such that the learning becomes unstable. The physics of FN-DAM ensures that weight decay (in the absence of any updates) towards a zero vector (due to resynchronization) which is important for neural network generalization [37].

Like other analog non-volatile memories, FN-DAM could be used in any previously proposed compute-in-memory (CIM) architectures. However, in conventional CIM implementations the weights are trained offline and then downloaded on chip without retraining the processor [38]. This makes the architecture prone to analog artifacts like offsets, mismatch and non-linearities. On-chip learning and training mitigates this problem whereby the weights self-calibrate for the artifacts to produce the desired output [39]. However, to support on-chip training/learning, weights need to be updated at a precision greater than 12 bits [34]. In this regard FN-DAM exhibit a significant advantage compared to other analog memories. Even though in this proof-of-concept work, we have a used a hybrid chip-in-the-loop training paradigm, it is anticipated that in the future the training circuits and FN-DAM modules could be integrated together on-chip.

## Methods

### Initialization of the FN-DAM array

For each node of each recorder, the readout voltage was programmed to around 3 V while the tunneling node was operating in the tunneling regime (Supplementary Fig. 1). This was achieved through a combination of tunneling and injection. Specifically, $V_{DD}$ was set to 7 V, input to 5 V, and the program tunneling pin was gradually increased to 23 V. Around 12–13V the tunneling node's potential would start increasing. The coupled readout node's potential would also increase. When the readout potential went over 4.5 V, electrons would start injecting into the readout floating gate, thus ensuring its potential was clamped below 5 V. After this initial programming, VDD was set to 6 V for the rest of the experiments. See Supplementary Section I for further details. After one-time programming, input was set to 0 V, input tunneling voltage was set to 21.5 V for 1 minute and then the floating gate was allowed to discharge naturally. Readout voltages for the SET and RESET nodes were measured every 500 milliseconds. The rate of discharge for each node was calculated; and a state where the tunneling rates would be equal was chosen as the initial synchronization point for the remainder of the experiments.

### FN Tunneling dynamics

V(t) is the floating gate voltage given by [22, 21]

$$V(t) = \frac{k_2}{\log(k_1 t + k_0)} \quad (1)$$

Where $k_1$ and $k_2$ are device specific parameters and $k_0$ depends on initial condition as:

$$k_0 = \exp\left(-\frac{k_2}{V_0}\right)$$

Using the dynamic given in Eqn.1, the Fowler-Nordheim tunneling current can be calculated as:

$$\frac{I_{FN}(V(t))}{C_T} = \frac{d(V(t))}{dt} = \left(\frac{k_1}{k_2}\right) V^2 \exp\left(-\frac{k_2}{V}\right) \quad (2)$$

### Weight decay model and FN-DAM dynamics

Many neural network training algorithms are based on solving an optimization problem of the form [26]:

$$\min_{\overline{w}} H(w) = \frac{\alpha}{2} \|\overline{w}\| + \mathcal{L}(\overline{w}) \quad (3)$$

where $\bar{w}$ denotes the network synaptic weights, $\mathcal{L}(\cdot)$ is a loss-function based on the training set and $\alpha$ is a hyper-parameter that controls the effect of the $\mathcal{L}_2$ regularization. Applying gradient descent updates on each element $w_i$ of the weight vector $\bar{w}$ as:

$$w_{i,n+1} - w_{i,n} = -\alpha \eta_n w_{i,n} - \eta_n \frac{\delta \mathcal{L}(\bar{w})}{\delta w_{i,n}} \tag{4}$$

Where the learning rate $\eta_n$ is chosen to vary according to $\eta_n \sim O(1/n)$ to ensure convergence to a local minimum [27]:

The naturally implemented weight decay dynamics in FN-DAM devices can be modeled by applying Kirchhoff's Current Law at the SET and RESET floating gate nodes (see Fig. 1e).

$$C_T \frac{d}{dt}(W_S) + I_{FN}(W_S) = C_C \frac{d}{dt}(V_{SET}) \tag{5}$$

$$C_T \frac{d}{dt}(W_R) + I_{FN}(W_R) = C_C \frac{d}{dt}(V_{RESET}) \tag{6}$$

Where $C_{FG} + C_C = C_T$ is the total capacitance at the floating gate. Taking the difference between the above two equations, we get:

$$C_T \frac{d}{dt}(W_S - W_R) + I_{FN}(W_S) - I_{FN}(W_R) = C_C \frac{d}{dt}(V_{SET} - V_{RESET}) \tag{7}$$

For the differential architecture, $w = W_S - W_R$. Let $V_{train} = V_{SET} - V_{RESET}$, the training voltage calculated by the training algorithm. In addition, $I_{FN}$ is substituted from Eqn. 2. Let $C_C/C_T = C_R$, the input coupling ratio:

$$\frac{dw}{dt} = -\frac{(I_{FN}(W_S) - I_{FN}(W_R))}{C_T} + C_R \frac{d}{dt}(V_{train}) \tag{8}$$

$$\frac{dw}{dt} = \frac{-\left(\frac{k_1}{k_2}\right) W_R^2 \exp\left(-\frac{k_2}{W_R}\right) + \left(\frac{k_1}{k_2}\right) W_S^2 \exp\left(-\frac{k_2}{W_S}\right)}{W_R - W_S} w + C_R \frac{d}{dt}(V_{train}) \tag{9}$$

Discretizing the update for a small time-interval $\Delta t$

$$w_{n+1} = w_n + \frac{-\left(\frac{k_1}{k_2}\right) W_R^2 \exp\left(-\frac{k_2}{W_R}\right) + \left(\frac{k_1}{k_2}\right) W_S^2 \exp\left(-\frac{k_2}{W_S}\right)}{W_R - W_S} w_n \Delta t + C_R \Delta V_{train,n} \tag{10}$$

Let $\mu = W_R/W_S$

$$w_{n+1} = w_n - \left(\frac{k_1}{k_2}\right) W_S \exp\left(-\frac{k_2}{W_S}\right) \frac{\mu^2 \exp\left(-\frac{k_2}{W_S}\left(1 - \frac{1}{\mu}\right)\right) - 1}{\mu - 1} w_n \Delta t + C_R \Delta V_{train,n} \tag{11}$$

Assuming that the stored weight (measured in mV) is much smaller than node potential (> 6V) i.e., $w \ll W_R$ (and $W_R \approx W_S$) and taking the limit ($\mu \to 1$) using L'Hôpital's rule:

$$w_{n+1} = \left(1 - \left(\frac{k_1}{k_2}\right)(2W_S + k_2)\exp\left(-\frac{k_2}{W_S}\right)\Delta t\right)w_n + C_R \Delta V_{train,n} \quad (12)$$

$W_S$ follows the temporal dynamics given in Eqn. 1,

$$w_{n+1} = -k_1\left(\frac{2}{\log(k_1 n\Delta t + k_0)} + 1\right)\left(\frac{1}{k_1 n\Delta t + k_0}\right)w_n \Delta t + C_R \Delta V_{train,n} \quad (13)$$

Comparing above equation to Eqn. 4, the weight decay factor for FN-DAM system is given as:

$$\alpha\eta_n = k_1\left(\frac{2}{\log(k_1 n\Delta t + k_0)} + 1\right)\left(\frac{1}{k_1 n\Delta t + k_0}\right) \to O\left(\frac{1}{n}\right) \quad (14)$$

### Chip-in-the-loop linear classifier training

A hybrid hardware-software system was implemented to carry out an online machine learning task. The physical weights ($\overline{w} = [w_1, w_2]$) stored in two FN-DAM devices were measured and used to classify points from a labelled test data set in software. We sought to train a linear decision boundary of the form:

$$f(\overline{x}, \overline{w}) = x_2 + w_1 x_1 + w_0 \quad (15)$$

$\overline{x} = [x_1, x_2]$ are the features of the training set. For each point that was misclassified, the error in the classification was calculated and a gradient of the loss function with respect to the weights was calculated. Based on the gradient information, the weights were updated in hardware by application of SET and RESET pulses via a function generator.

The states of the SET and RESET nodes were measured every 2 seconds and the weight of each memory cell, $i$, was calculated as:

$$w_i = 1000 * (W_{R,i} - W_{S,i}) \quad (16)$$

The factor of 1000 indicates that the weight is stored as the potential difference between the SET and RESET nodes as measured in mV. We followed a stochastic gradient descent method. We defined loss function as:

$$\mathcal{L}_n(\overline{w}) = ReLU\left(1 - y_n f(\overline{x_n}, \overline{w})\right) \quad (17)$$

The gradient of the loss function was calculated as:

$$G_n(\overline{w}) = \frac{\partial \mathcal{L}_n(\overline{w})}{\partial \overline{w}} \quad (18)$$

The weights needed to be updated as

$$w_{n+1} = w_n - \lambda_n G_n(\overline{w}) \quad (19)$$

Here $\lambda_n$ is the learning rate as set by the learning algorithm. The gradient information is used to update FN-DAM by applying control pulses to SET/RESET nodes via a suitable mapping function $T$:

$$V_{train,n} = T(\lambda_n G_n(\overline{w})) \quad (20)$$

Positive weight updates were carried out by application of SET pulses and negative updates via RESET pulses. The magnitude of the update was implemented by modulating the number of input pulses.

# Supplementary Information

## I. Initial Programming

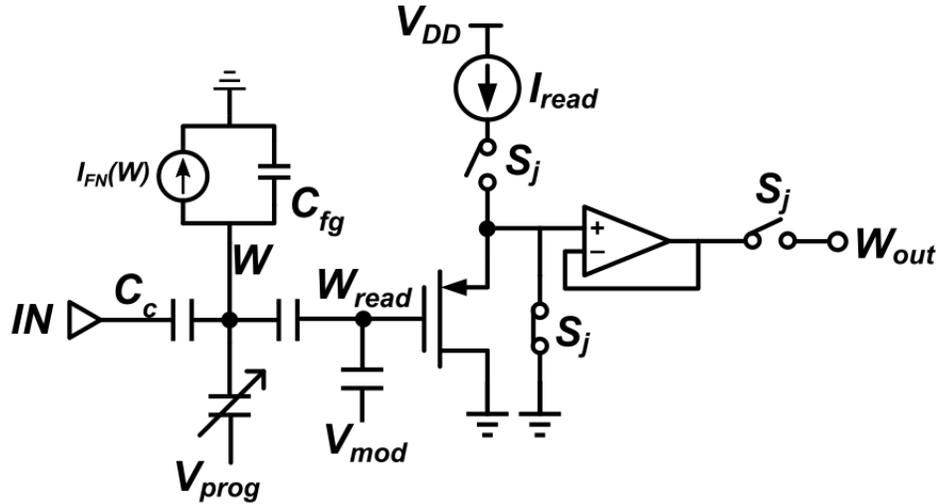

*Figure 1: Architecture of one half of the FN-DAM, which can be configured as a SET or a RESET node.*

The FN-DAM consists of two identical nodes: SET and RESET node. Each node contains two floating gates decoupled via a capacitor (SI Figure 1) – tunneling gate (W) and readout gate ($W_{read}$). The charge on each gate of the system can be individually programmed using a combination of tunneling (to increases charge, coarse tuning) and hot electron injection (to decreases charge, fine tuning). The tunneling gate, which stores the dynamic analog memory, is biased in the FN tunneling regime. By setting $V_{prog}$ to a high potential of 22 V, the tunneling node is set to ~8 V which is sufficient to initiate observable FN tunneling for oxide thickness of around 13 nm. $W_{read}$ is capacitively decoupled from the tunneling node to avert readout disturbances. The readout node is biased at a lower voltage to prevent injection into the readout node during operation. The potential of the readout node is lowered through hot electron injection. Injection is initiated by setting $V_{DD}$= 7 V and the input pin to a value such that $V_{DS}$ across the PMOS is above 4.2 V. Switch $S_j$ allows for individual control of each FN-DAM block for reading and programming.

## II.   Write Energy Dissipation

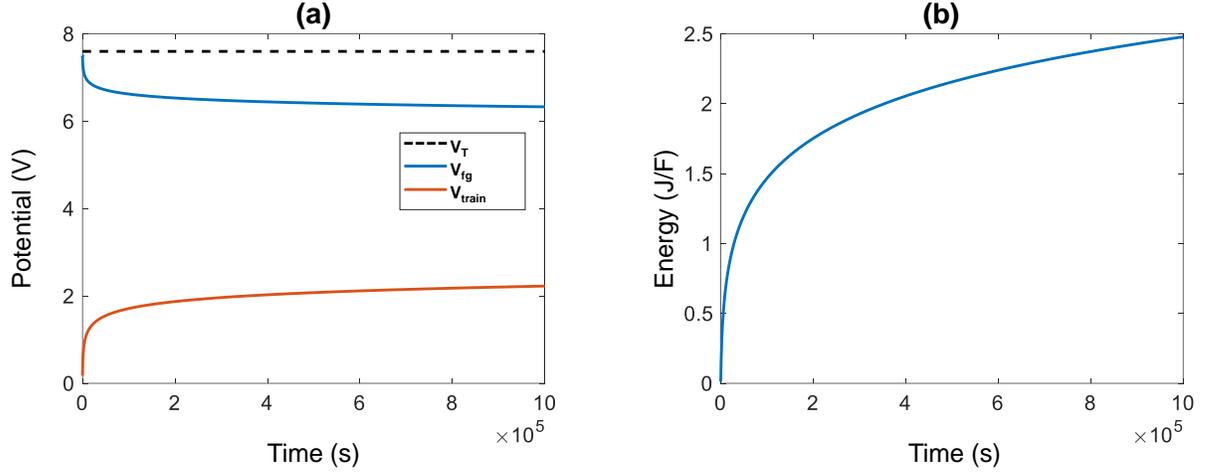

*Figure 2: a) Target voltage, floating gate voltage and training voltage as a function of time. b) Energy required to charge unit capacitance as a function of time.*

The magnitude of input pulse required, $V_{train}(t)$ (Fig. 2a) so that the floating gate node at current potential $V_{FG}(t)$ shifts to a target voltage $V_T$ is given by:

$$V_{train}(t) = \frac{V_T - V_{FG}(t)}{C_R}$$

Where $C_R$ is the input capacitive coupling ratio $C_R = \frac{C_C}{C_C + C_{FG}}$. The floating gate voltage $V_{FG}(t)$ is approximated by the following dynamic [1]:

$$V_{FG}(t) = \frac{k_2}{\log(k_1 t + k_0)} \qquad (1)$$

The energy required to charge the input capacitor is given as

$$E(t) = \frac{1}{2} C_{in} (V_{in}(t))^2$$

Figure 2b shows instantaneous energy required to charge unit capacitance when $V_T = 7.6V$ and $V_{FG}(0) = 7.5V$. The input capacitance of our device was 1 pF, and the instantaneous write energy per update increased from 5 fJ to 2.5pJ over 12 days.

# III. Memory Retention

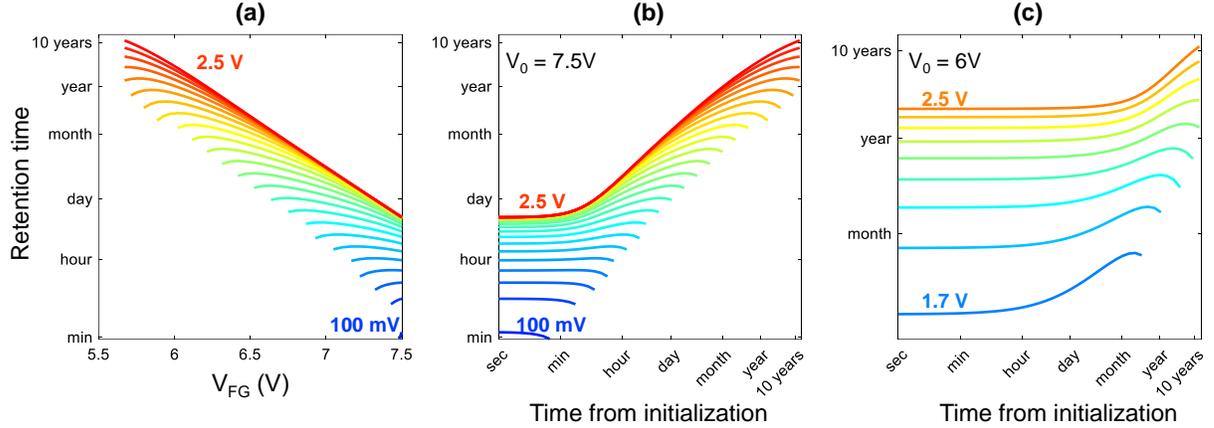

*Figure 3: a) Retention time as a function of floating gate voltage for a range of input pulse magnitudes. b-c) Retention time of weight updates as a function of time elapsed after initialization to 7.5V (b) and 6V (c).*

The method for calculating retention time of dynamical systems was described in [2]. In brief, the point at which the analog memory – due to resynchronization – falls below the noise floor is the retention time. The noise floor consists of a constant noise introduced by the readout noise and an operational noise that increases with time, due to thermally induced random desynchronization.

Total noise: $\sigma_T(t) = \sigma_0 + \sigma(t)$

In this case, we assumed $\sigma_0 = 100\ \mu V$ and estimated $\sigma_t = 1.4 t^{0.5}\ \mu V$ from experiments without any external pulse.

At $T_{Ret}$, synaptic memory's state goes below the noise floor and hence the following condition is satisfied:

$$w(T_{Ret}) = \sigma_T(T_{Ret})$$

When the FN-DAM is biased at around 6 V, its retention time is similar to FLASH/EEPROM memory. However, energy consumption is around $150\ fJ$ (for a $100\ fF$ input capacitance).

# IV. Deep neural network simulations

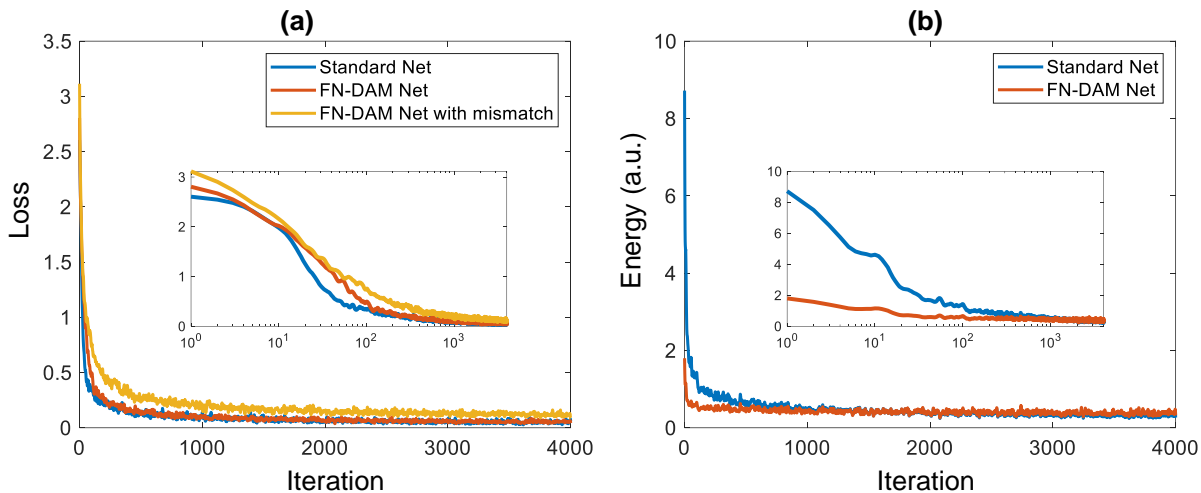

*Figure 4: a) Network loss for 3 types of network models. Inset shows same data with X axis in log scale. b) Energy spent in updating the network weights. Inset shows same data with X axis in log scale.*

The performance of FG-DAM model was compared to that of a standard network model. A 15-layer convolutional neural network was trained on the MNIST dataset using the MATLAB Deep Learning Toolbox. For each learnable parameter in the CNN, a software FN-DAM instance corresponding to that parameter was created. In each iteration, the loss of the network function and gradients were calculated. The gradients were used to update the weights via Stochastic Gradient Descent with Momentum (SGDM) algorithm. The updated weights were mapped onto the FN-DAM array. The weights in the FN-DAM array were decayed according to Eqn. 14. These weights were then mapped back into the CNN. This learning process was carried on for 9 epochs. In the 10$^{th}$ epoch, no gradient updates were performed. However, the weights were allowed to decay for the last epoch (note that in the standard CNN case, the memory was static). A special case with a 0.1% randomly assigned mismatch in the floating gate parameters ($k_1$ and $k_2$) was also implemented.

*Table 1: Comparison metrics*

| Accuracy (%) | After 9 epochs | After 10 epochs |
|---|---|---|
| Standard CNN | 98.6 | 98.6 |
| FN-DAM CNN | 99 | 99.2 |
| FN-DAM CNN with mismatch | 97.4 | 96.3 |

## V. Read Energy Dissipation

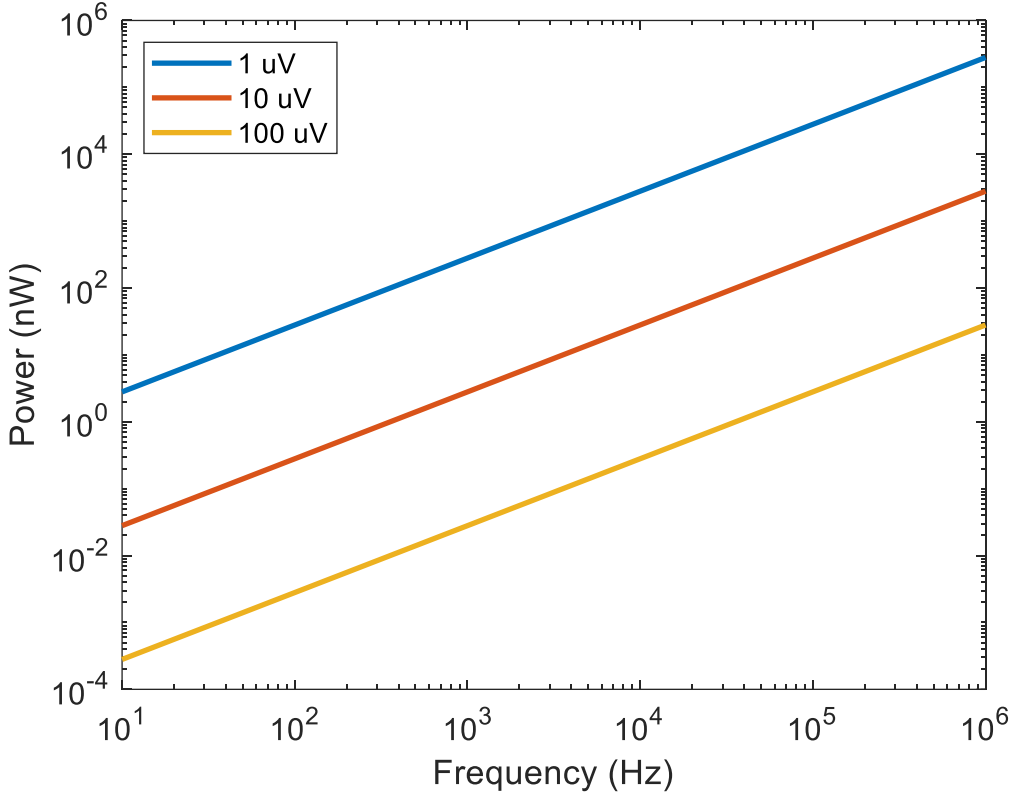

*Figure 5: Minimum power required to read floating gate voltage as a function of required readout speed. Noise floors shown in legend.*

The readout power is dependent on the readout accuracy required and the speed at which it operates.

For a PMOS in a source follower configuration, the readout noise is given by:

$$V_n^2 = \frac{4kT}{g_m}\Delta f = \frac{4kT}{q}*\frac{q}{g_m}\Delta f = \frac{4U_T q}{g_m}\Delta f$$

For subthreshold operation,

$$g_m = \frac{\kappa I_d}{U_T}$$

$$\therefore V_n^2 = \frac{4U_T^2 q}{\kappa I_d}\Delta f = \frac{4U_T^2 q V_{DD}}{\kappa P_{read}}\Delta f$$

Above equation is plotted in SI Figure 5 for different noise floors and readout frequency for $V_{dd} = 5V$, $U_T = 26\ mV$ and $\kappa = 0.7$

# VI. Programming dynamics

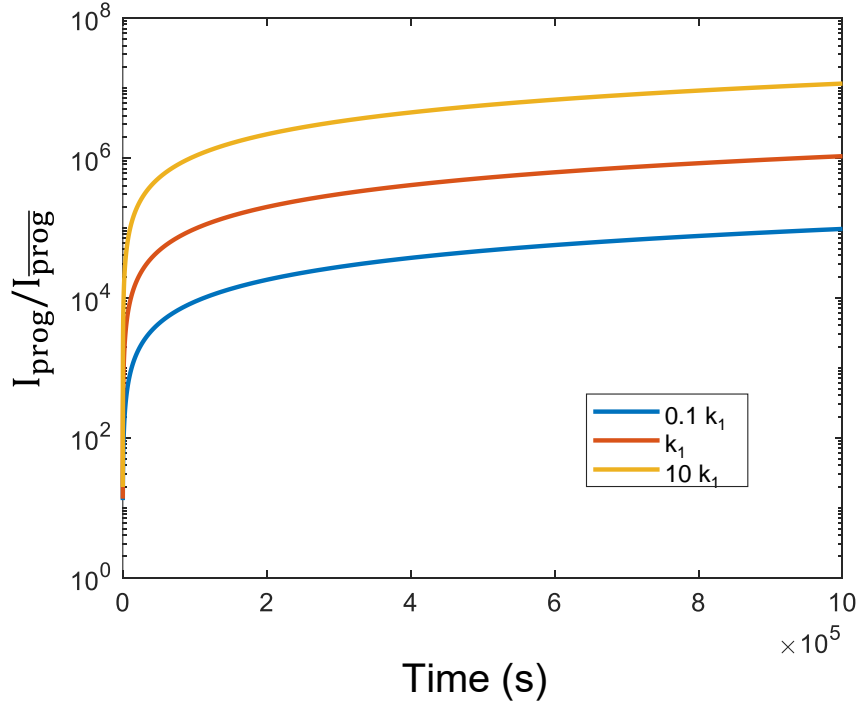

*Figure 6: Programming ratio for different $k_1$ parameter which can be controlled by changing the size of tunneling junction.*

The FN-DAM is programmed by applying a pulse of magnitude $V_{train}(t)$ so that the node reaches a potential of $V_T$ through the input coupling capacitor, as derived in the previous section. The programming ratio is given by:

$$\frac{I_{prog}}{I_{\overline{prog}}} = \frac{I_{FN}(V_T)}{I_{FN}(V_{FG}(t))}$$

Dynamics of FN tunneling current are given by [1]:

$$\frac{I_{FN}(V(t))}{C_T} = \frac{d(V(t))}{dt} = \left(\frac{k_1}{k_2}\right) V^2 \exp\left(-\frac{k_2}{V}\right)$$

$$\frac{I_{prog}}{I_{\overline{prog}}} = \left(\frac{V_T}{V_{FG}(t)}\right)^2 \exp\left(\frac{k_2}{V_{FG}(t)} - \frac{k_2}{V_T}\right)$$

Above equation is plotted for 3 values of $k_1$ which affect the dynamics of $V_{FG}(t)$. The parameter $k_1$ can be altered during the design phase by changing the area and capacitance of the floating gate node.

# Supplementary References